\documentclass[letterpaper, 10 pt, journal, twoside]{IEEEtran}
\usepackage{amsmath,amssymb,amsfonts}
\usepackage{array}
\usepackage[caption=false,font=normalsize,labelfont=sf,textfont=sf]{subfig}
\usepackage{textcomp}
\usepackage{stfloats}
\usepackage{url}
\usepackage{verbatim}
\usepackage{graphicx}
\usepackage{cite}
\usepackage{xcolor}
\usepackage{multirow}
\usepackage{algorithm}
\usepackage{algpseudocode}
\definecolor{redtrans}{RGB}{255,179,179}
\usepackage[hidelinks]{hyperref}
\usepackage{orcidlink}
\begin{document}
\bstctlcite{IEEEexample:BSTcontrol}

\title{SEDMamba: Enhancing  Selective State Space Modelling with Bottleneck Mechanism and Fine-to-Coarse Temporal Fusion for Efficient Error Detection in Robot-Assisted Surgery}

\author{Jialang Xu$^{1}$\orcidlink{0000-0003-2324-7033}, Nazir Sirajudeen$^{1}$\orcidlink{0009-0006-4221-1828}, Matthew Boal$^{2}$\orcidlink{0000-0002-7288-3354}, Nader Francis$^{2}$\orcidlink{0000-0001-8498-9175}, \\
Danail Stoyanov$^{1}$\orcidlink{0000-0002-0980-3227}, \IEEEmembership{Fellow, IEEE}, and Evangelos B. Mazomenos$^{1}$\orcidlink{0000-0003-0357-5996}, \IEEEmembership{Member, IEEE}%
\thanks{Received 17 September 2024; accepted 6 November 2024. Date of publication; date of current version. This article was recommended for publication by Associate Editor Jie Ying Wu and Editor Pietro Valdastri upon evaluation of the reviewers’ comments. This work was supported in part by the Wellcome/EPSRC Centre for Interventional and Surgical Sciences (WEISS) under Grant 203145Z/16/Z and Grant NS/A000050/1, in part by the EPSRC-funded UCL Centre for Doctoral Training in Intelligent, Integrated Imaging in Healthcare (i4health) under Grant EP/S021930/1, in part by the UCL Research Excellence Scholarship, in part by the Department of Science, Innovation and Technology (DSIT), and in part by the Royal Academy of Engineering under the Chair in Emerging Technologies Programme. (\textit{Corresponding authors: Jialang Xu; Evangelos B. Mazomenos.})}%
\thanks{$^{1}$Jialang Xu, Nazir Sirajudeen, Danail Stoyanov, and Evangelos B. Mazomenos are with the UCL Hawkes Institute and Department of Medical Physics and Biomedical Engineering, University College London, WC1E 6BT London, U.K. {\tt\footnotesize\{jialang.xu.22; nazir.sirajudeen.20; danail.stoyanov; e.mazomenos\}@ucl.ac.uk}}%
\thanks{$^{2}$Matthew Boal and Nader Francis are with the Griffin Institute, Northwick Park and St Mark’s Hospital, HA1 3UJ Harrow, U.K., and also with the University College London, WC1E 6BT London, U.K. {\tt\footnotesize\{m.boal; n.francis\}@griffininstitute.org.uk}}%
\thanks{For the purpose of open access, the author has applied a Creative Commons Attribution (CC
BY) license to any Author Accepted Manuscript version arising.}
\thanks{Digital Object Identifier (DOI): 10.1109/LRA.2024.3505818.}

}

\markboth{IEEE Robotics and Automation Letters. Preprint Version. Accepted November, 2024}
{Xu \MakeLowercase{\textit{et al.}}: SEDM\MakeLowercase{amba}} 

\IEEEpubid{~\copyright~2024 IEEE. Personal use is permitted, but republication/redistribution requires IEEE permission. See https://www.ieee.org/publications/rights/index.html for more information}

\maketitle

\begin{abstract}
Automated detection of surgical errors can improve robotic-assisted surgery. Despite promising progress, existing methods still face challenges in capturing rich temporal context to establish long-term dependencies while maintaining computational efficiency. In this paper, we propose a novel hierarchical model named SEDMamba, which incorporates the selective state space model (SSM) into surgical error detection, facilitating efficient long sequence modelling with linear complexity. SEDMamba enhances selective SSM with a bottleneck mechanism and fine-to-coarse temporal fusion (FCTF) to detect and temporally localize surgical errors in long videos. The bottleneck mechanism compresses and restores features within their spatial dimension, thereby reducing computational complexity. FCTF utilizes multiple dilated 1D convolutional layers to merge temporal information across diverse scale ranges, accommodating errors of varying duration. Our work also contributes the first-of-its-kind, frame-level, in-vivo surgical error dataset to support error detection in real surgical cases. Specifically, we deploy the clinically validated observational clinical human reliability assessment tool (OCHRA) to annotate the errors during suturing tasks in an open-source radical prostatectomy dataset (SAR-RARP50). Experimental results demonstrate that our SEDMamba outperforms state-of-the-art methods with at least 1.82\% AUC and 3.80\% AP performance gains with significantly reduced computational complexity. The corresponding error annotations, code and models are released at https://github.com/wzjialang/SEDMamba.
\end{abstract}


\section{Introduction}
\IEEEPARstart{R}{obotic-assisted} surgery (RAS) is well-adopted globally and the preferred option for many surgical specialties (e.g. urology)~\cite{robot_rarp}. RAS is complex, highly variable and requires advanced technical and cognitive skills. Adverse events during RAS and broadly laparoscopic surgery are not uncommon~\cite{sarker_errors}. A proportion of these are operator's errors leading to consequences ranging from suboptimal execution of surgical tasks to patient injury and rarely death. Overall, human errors are the second cause of injury and death in RAS~\cite{alemzadeh_adverse}. 

Observational clinical human reliability assessment (OCHRA) is a standardized method to assess the quality of surgical execution by detecting and characterising the nature (procedural or executional) and severity of technical errors on various specialties~\cite{tang_ochra,gorard_ochra,eubanks_score}. Clinical validation has identified executional errors, following the OCRHA definitions, as a strong predictor of both skill and patient outcomes~\cite{curtis_eval,curtis_assoc}. This strongly promotes the application of such tools for reducing adverse events and surgical training. Deployment of OCHRA is impractical due to the significant time required for reviewing and manual annotation. Subjectivity in validity and reliability issues further prioritize the need for automated solutions~\cite{boal_review,sds}. 

\IEEEpubidadjcol

Propelled by deep learning (DL) methods and the public JIGSAWS dataset~\cite{jigsaws}, containing atomic gesture and skill annotations from RAS tasks in dry labs, significant prior work on RAS video understanding for skill analysis focused on recognition and temporal localization of surgical gestures~\cite{funke_3d,zhang_symm,liu_deep,dipietro_reco,zhang_sdnet}. Recent work adapted DL methods for gesture recognition to in-vivo datasets\cite{act_review,rarp,kiyasseh_vis}, and also studied the association of the observed gesture sequence to patient outcomes and surgeon's skill~\cite{ma_gest}. Detection and temporal localization of procedural and executional errors have not been extensively studied. Hutchinson et al.~\cite{jigsaws_errors} incorporated error annotations in two JIGSAWS tasks (simulation suturing and needle passing). Procedural errors are considered when a surgical gesture is omitted or the task is carried out with a gesture sequence that deviates from the defined grammar graph, while executional errors denote moments of suboptimal performance. For each gesture, a maximum of four executional errors were considered from a list (multiple attempts, needle orientation, needle drop, out of view), used to characterize the entire (begin to end) gesture as erroneous (if any error was noted) or normal~\cite{jigsaws_errors}. Employing these annotations, Li et. al. compared four DL classifiers for detecting executional errors on gesture-level clips~\cite{run_detection}. We argue that the annotation approach in~\cite{jigsaws_errors}, though successful in dry-lab, is not directly applicable to in-vivo RAS settings. It is typical for errors to span across multiple gestures, thus their temporal localization can not always align to the temporal boundaries (begin, end) of gestures. Consequently, its annotation based on pre-segmented gesture clips fails to accurately provide temporal boundaries for the start and end of errors. In this work, we define custom error descriptions, following OCHRA, and present a novel in-vivo RAS dataset with first-of-its-kind, frame-level error annotations to provide more accurate temporal boundaries of errors for surgical error detection (SED).

SED aims to detect errors that occur during surgery. It has been formulated as a binary classification task, where the ground truth labels are `\textit{normal}': if no error exists, or `\textit{error}': if any error type exists~\cite{run_detection,drautoencoder}. In~\cite{run_detection}, kinematic data was downsampled to segments of 30 data points, using sliding windows within manually pre-segmented gesture clips, serving as the input for a long short-term memory (LSTM) network to classify each gesture clip as either error or normal. However, its parallel computing capability is constrained by the inherently state-dependency nature of LSTM. Samuel and Cuzzolin~\cite{drautoencoder} developed an unsupervised convolutional autoencoder to categorise each frame as either error or normal in pre-trimmed video clips, yet the local receptive field of convolution layers hinders its global temporal modelling capability. Overall, existing research on SED has primarily focused on short kinematics data and video clips, efficient SED in long surgical videos remains unexplored. To achieve this objective, it is essential to consider two key factors: long-term dependency modelling and computational complexity. Transformer architectures, benefiting from the self-attention mechanism, enable global modelling and have demonstrated impressive performance in surgical video analysis tasks such as surgical phase recognition~\cite{asformer,transsv}. Nevertheless, the quadratic computational complexity of self-attention with respect to video sequence length significantly increases the computational burden, impeding its efficiency in handling long video scenarios. 

\begin{figure*}[!tbh]
    \centering
    \includegraphics[width=0.84\textwidth]{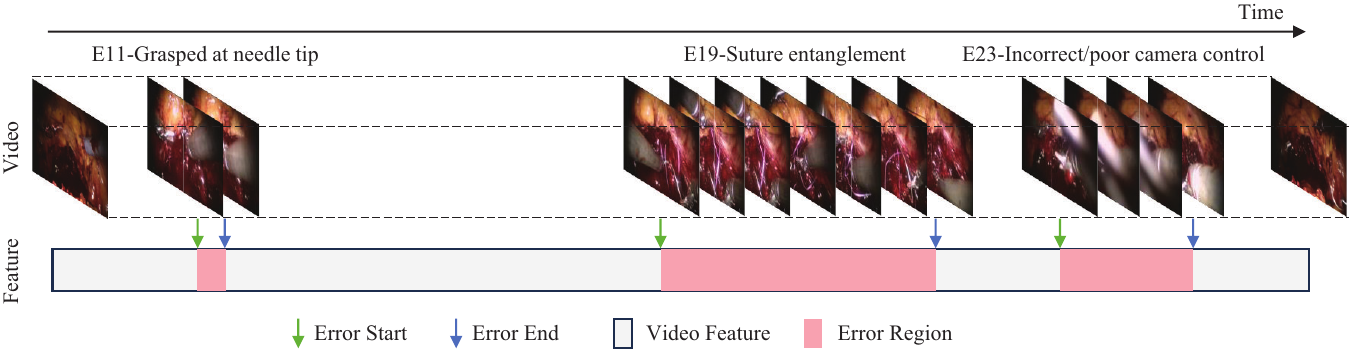}
    \caption{A surgical video from the SAR-RARP50 dataset, including errors of varying duration.}
    \label{fig_errorduration}
\end{figure*}

Recently, state space models (SSMs) have proven highly effective in capturing long sequences in the natural language processing domain~\cite{s4,mamba}. Selective SSM~\cite{mamba}, namely Mamba, distinguishes itself from other SSM-based methods by integrating a data-dependent SSM with a selection mechanism that employs parallel scanning, thereby achieving linear complexity while preserving global receptive fields. Inspired by the great success of Mamba, researchers have explored it in the computer vision domain, yielding promising performance. Zhu et al.~\cite{zhu2024vision} proposed a bidirectional Mamba (Vim) that scans images forward and backward to enhance spatial-aware understanding for image tasks, significantly improving computational efficiency while achieving higher performance compared to well-established Transformers. SegMamba~\cite{xing2024segmamba} constructed a tri-oriented Mamba for 3D medical image segmentation. However, as depicted in Fig.~\ref{fig_errorduration}, surgical videos contain errors with varied duration, necessitating the algorithm to consider temporal information at different granularities.  Prior selective SSM-based works rarely consider modelling multi-scale temporal information, we thus showcase the effectiveness of a novel selective SSM-based method named SEDMamba, which captures temporal information across various scales to detect surgical errors in suturing from robotic-assisted radical prostatectomy (RARP) cases.

This work makes the following contributions: 

1) We propose a novel hierarchical model, SEDMamba, constructed with bottleneck multi-scale state space (BMSS) blocks to detect surgical errors in long videos from real RAS cases. To handle errors with variable duration, we design a fine-to-coarse temporal fusion (FCTF) using different dilated 1D convolutional layers to merge temporal information across different scale ranges. Besides, to further reduce computational complexity, we incorporate a bottleneck mechanism that compresses the spatial dimension of features, enabling selective SSM to concentrate on modelling the temporal dimension. To the best of our knowledge, this is the first attempt to exploit selective SSM in SED. 

2) We present the first-of-its-kind, frame-level surgical error annotations of an in-vivo RAS dataset and apply them to train and validate the proposed SEDMamba model. Our work establishes a roadmap of how to utilize this new dataset for developing SED benchmarks.

3) The proposed SEDMamba outperforms state-of-the-art methods by 1.82-13.06\% AUC and 3.8-18.86\% AP while requiring 11.71-91.67\% fewer model parameters and 40.71-92.32\% fewer FLOPs.

\begin{figure}[!tbh]
    \centering
    \includegraphics[width=0.8\columnwidth]{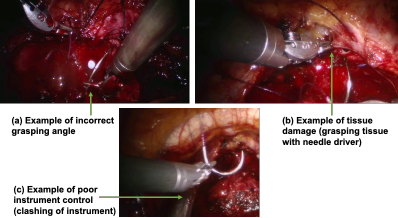}
    \caption{Example error annotations at SAR-RARP50: (a) E6-Incorrect angle grasping needle; (b) E5-Tissue damage; (c) E24-Poor instrument control (clashing).}
    \label{error_samples}
\end{figure}

\begin{table}[!tbh]
\centering
\caption{Error coding descriptions used for annotation}
\label{table_ecd}
\resizebox{\columnwidth}{!}{
\begin{tabular}{|c|l|c|c|}
\hline
Error Type & Error Description & Incidence &Total Frames        \\ \hline
E1    & \begin{tabular}[c]{@{}l@{}} Multiple attempts (end of first attempt \\ to end of successful attempt) i.e. multiple piercings,\\ not regrasping with needle held in tissue \end{tabular}    & 510  & 119635\\  \hline
E2   & Needle drop/slip if in tissue      & 77  & 3673\\ \hline
E3   & Instrument(s) out of view          & 1   & 27    \\ \hline
E4   & Needle out of view                  & 259   & 32548   \\ \hline
E5   & Tissue damage including poor/error in tissue stabilisation  & 113     & 16799 \\ \hline
E6   & Incorrect angle grasping needle (not perpendicular) & 116     & 2237   \\ \hline
E7   & Incorrect position along needle (normally second/third)   & 89    & 1737       \\ \hline
E8   & Excessive force                          & 65    & 6220   \\ \hline
E9   & Needle does not follow the curve       &169      &15078                    \\ \hline
E10  & Needle entry incorrect angle i.e. not perpendicular  & 35    &2761      \\ \hline
E11  & Grasped at needle tip               &106     &888   \\ \hline
E12  & Suture is loosened                      &0   &0              \\ \hline
E13  &   Thread caught in instrument              &67   &14423     \\ \hline
E14  & Knot tied is not square (C/Reverse C)        &0  &0  \\ \hline
E15  & Inadequate number of throws                    &1    &93            \\ \hline
E16  & Suture pulled through tissue before tying knot   &0  &0          \\ \hline
E17  & Incorrect distancing between needle drives (too close/far)   &4  &977 \\ \hline
E18  & Suture not pulled through between needle drives     &4   &886       \\ \hline
E19  & Suture entanglement                       &34    &20364           \\ \hline
E20  & Fraying the suture                        &5 &600      \\ \hline
E21  & Snapping the suture                        &4    &67     \\ \hline
E22  & Dangerous/poor/incorrect needle disposal     &0  &0          \\ \hline
E23  & Incorrect/poor camera control (blurred/poor view) &162   &34204      \\ \hline
E24  &  \begin{tabular}[c]{@{}l@{}} Incorrect/poor instrument control (clashing, \\ 3rd arm/non-dominant use) \end{tabular} & 686   &47138 \\ \hline
\end{tabular}
}
\end{table}

\section{Methodology}
\subsection{Dataset Description and Annotation}
A total of 48 videos, sampled at 60 Hz, from the open-source, SAR-RARP50~\cite{sarrarp50,rarp45}, dataset of RARP cases were independently and blindly reviewed by two assessors. SAR-RARP50 focuses on the suturing of the dorsal venous complex (DVC) task and contains open-source annotations of eight gestures used as a task analysis sheet~\cite{sarrarp50}. We developed a customized tool for the annotation of both procedural and executional errors using a previously validated suturing checklist and the OCHRA methodology~\cite{guni_check}. This was summarized as a list of possible procedural and executional errors within any given suturing task, as seen in Table~\ref{table_ecd}. The duration of these videos varies from one minute to over eleven minutes, including the complete suturing process. Two assessors (one senior and one junior RAS surgeon with expertise in RAS education) analyzed the untrimmed videos sequentially frame-by-frame and annotated the start and end frame indexes of all errors. Fig.~\ref{error_samples} illustrates three types of errors from our annotations. Inter-rater reliability was verified with Pearson's correlation showing very strong agreement ($r$=0.8726, $p<$0.001) between the two assessors. Bland-Altman analysis shows average of 46.43 errors/case, and average difference of 5.87 errors between raters. Consensus was established for all disagreements via joint review.  

Overall, 2507 error events were annotated, with E24-``Incorrect/poor instrument control'' occurring the most frequently (686), followed by E1-``Multiple attempts'' (510). The frequency of error types had statistically significant difference (Chi-squared test: $\chi2$ = 1945.15, $p<$ 2.2e-16). The duration of errors varies, with the longest average duration being E19-``Suture entanglement", averaging around 599 frames per case, while the shortest is E11-``Grasped at needle tip", averaging 8 frames per case. The 48 videos comprise a total of 976037 frames. In the SED task, the frame is categorized as `\textit{normal}' if no error exists or `\textit{error}' if any type of error exists, resulting in 320355 error frames. To detect errors of varying duration, irrespective of their type, it is crucial to model temporal information across multiple scales.

\subsection{Surgical Error Detection Model (SEDMamba)}
This section introduces the preliminary concepts related to SEDMamba, including state space modelling and discretization. We then formulate the SEDMamba architecture and its core bottleneck multi-scale state space (BMSS) block.

\begin{figure*}[!t]
    \centering
    \includegraphics[width=0.9\textwidth]{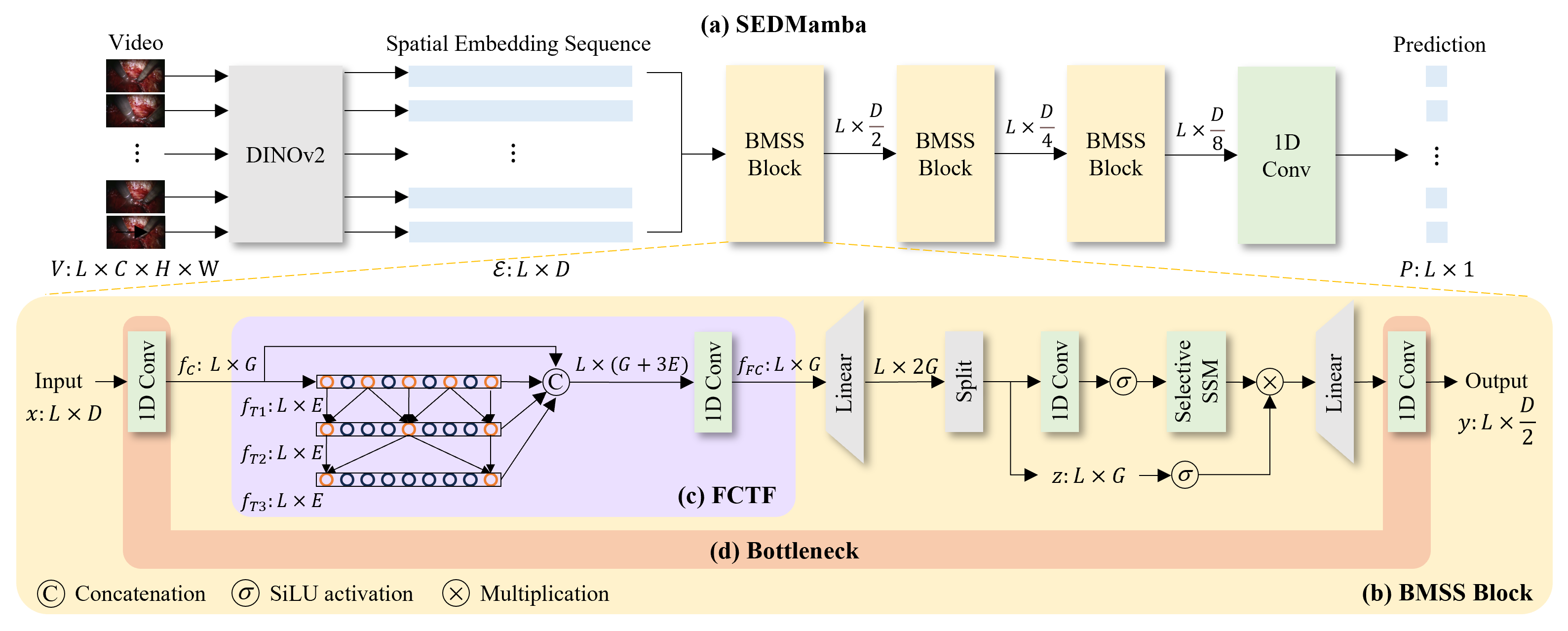}
    \caption{The pipeline of the proposed SEDMamba. (a) The overall architecture of SEDMamba. (b) The fundamental block of SEDMamba, namely the bottleneck multi-scale state space (BMSS) block; (c) Fine-to-coarse temporal fusion (FCTF); (d) Bottleneck mechanism.}
    \label{fig_arc}
\end{figure*}

\subsubsection{Preliminaries}
State space models (SSMs), such as structured state space sequence models~\cite{s4} and selective SSM~\cite{mamba}, are inspired by continuous system that map 1D feature or sequence $x(t) \in \mathbb{R}^L$ with length $L$ to $y(t) \in \mathbb{R}^L$ through a hidden state $h(t) \in \mathbb{R}^N$. Formally, SSMs can be formulated as:
\begin{equation}
\label{eq_ode}
h^{\prime}(t)=\mathbf{A} h(t)+\mathbf{B} x(t); y(t)=\mathbf{C} h(t)
\end{equation}
where $\mathbf{A} \in \mathbb{R}^{N \times N}$ is the evolution parameter, $\mathbf{B},\mathbf{C} \in \mathbb{R}^{N \times 1}$ denote as the projection parameters, $N$ is the state size. To transform the continuous parameters $\mathbf{A}, \mathbf{B}$ in Eq.~\ref{eq_ode} to discrete $\overline{\mathbf{A}}, \overline{\mathbf{B}}$, SSMs generally adopt the zero-order hold rule including a timescale parameter $\Delta$ for discretization:
\begin{equation}
\label{eq_disc}
\overline{\mathbf{A}}=\exp (\Delta \mathbf{A}), \overline{\mathbf{B}}=(\Delta \mathbf{A})^{-1}(\exp (\Delta \mathbf{A})-\mathbf{I}) \cdot \Delta \mathbf{B}
\end{equation}
Eq.~\ref{eq_ode} can now be rewritten as a discretized version based on Eq.~\ref{eq_disc}:
\begin{equation}
h_t=\overline{\mathbf{A}} h_{t-1}+\overline{\mathbf{B}} x_t; y_t=\mathbf{C} h_t
\end{equation}
The output $y$ of the model can be computed through a global convolution for efficient training:
\begin{equation}
\overline{\mathbf{K}} =\left(\mathbf{C} \overline{\mathbf{B}}, \mathbf{C} \overline{\mathbf{A B}}, \ldots, \mathbf{C} \overline{\mathbf{A}}^L \overline{\mathbf{B}}, \ldots\right); y =x * \overline{\mathbf{K}}
\end{equation}
where $L$ is the length of the input sequence $x$, and $\overline{\mathrm{K}} \in \mathbb{R}^L$ is a structured convolutional kernel~\cite{s4}. 

\subsubsection{Overview of SEDMamba}
SEDMamba has a hierarchical architecture, shown in Fig.~\ref{fig_arc}, comprising three BMSS blocks. Taking a video $V \in \mathbb{R}^{L\times C\times H\times W}$ with $L$ length, we first extract the spatial embedding sequence $\mathcal{E} \in \mathbb{R}^{L\times D}$ from a fixed DINOv2~\cite{dinov2}, where $D$ is the spatial dimension. DINOv2 is a strong vision foundation model proven to extract task-agnostic visual spatial embedding and excel in video-based tasks~\cite{dinov2}. Then, SEDMamba compresses $\mathcal{E}$ in the spatial dimension and captures fine-to-coarse temporal feature through each BMSS block, resulting in $i$-th hierarchical spatial-temporal representations with size $L\times \frac{D}{2^i}$. The 1D convolutional layer with 1 kernel size and 1 channel is then used for binary classification and outputs the final error probability prediction $P \in \mathbb{R}^{L\times 1}$.  Note that each value in $P$ indicates the likelihood that the corresponding frame is erroneous. We use binary cross-entropy loss $\mathcal{L}_{bce}$ for supervised training.

\subsubsection{Bottleneck Multi-scale State Space (BMSS) Block}
The structure of the BMSS block is shown in Fig.~\ref{fig_arc}(a). Encouraged by the strong long-range modelling ability of SSMs, the BMSS block incorporates selective SSM with bottleneck mechanism and fine-to-coarse temporal fusion (FCTF) to reduce computational complexity and capture inter-frame long-range dependencies across multiple temporal scales based on spatial embedding sequences, respectively. Given input $x \in \mathbb{R}^{L\times D}$, it first undergoes a bottleneck 1D convolution to compress its spatial dimension from $D$ to $G$, where compression factor $G\ll D$. Multi-scale temporal information is then obtained and merged via FCTF across four different temporal scales to generate fine-to-coarse temporal feature $f_{FC}$. After that, the $f_{FC}$ is expanded with a linear layer and split into two information flows. The first flow passes through a 1D convolutional layer, followed by a SiLU activation function~\cite{elfwing2018sigmoid} before entering the selective SSM. The output of the selective SSM is then gated by the $z$ generated from the other flow. Finally, we use another bottleneck 1D convolutional layer to restore the spatial dimension to $\frac{D}{2}$ and get the output of the BMSS block $y \in \mathbb{R}^{L\times \frac{D}{2}}$. \footnote{Pseudo-code of the BMSS block is provided as Algorithm 1 in the supplementary video.} The BMSS block inherits the linear complexity of the selective SSM while retaining a local-to-global receptive field, which is promising for video-based temporal error detection. 

\subsubsection{Fine-to-Coarse Temporal Fusion (FCTF)}
To tackle varying duration of error and restricted receptive fields of convolutional layers, we propose the FCTF to capture and merge multi-scale temporal information. FCTF stacks three dilated 1D convolution layers with dilation rates of 2, 4, and 8 from top to bottom, enabling the fine-to-coarse temporal feature extraction scale by scale. Besides, stacking dilated convolution increases the receptive field without the need to increase the parameter number by increasing the kernel size. The receptive field expands exponentially as the number of stacked layers increases. Therefore, with a few parameters, we achieve a significantly large receptive field in the temporal dimension, which mitigates model overfitting and facilitates efficient error detection. The receptive field at each layer, with a kernel size of 3, can be calculated with the following formula:
\begin{equation}
\text {ReceptiveField}(l)=2^{l+2}-1
\end{equation}
where $l\in[1,3]$ represents the layer number. 

Specifically, given the compressed feature $f_C\in\mathbb{R}^{L\times G}$, which is obtained via the first bottleneck 1D convolution layer. It first goes through a 1D convolution layer with 2 dilated rate, which extracts a temporal feature of a receptive field size of 7, denoted as $f_{T1}$. Similarly, the $f_{T1}$ undergoes further refinement through dilated 1D convolution layers with 4 and 8 dilation rates, thereby acquiring temporal features $f_{T2}$ and $f_{T3}$ across larger and coarser time spans. These temporal features of varying granularity are concatenated with the original scale $f_C$ along the spatial dimension, and ultimately, a 1D convolution with 1 kernel size fuse them into a fine-to-coarse temporal feature $f_{FC}\in\mathbb{R}^{L\times G}$.

\section{Experiments}
\subsection{Experimental Settings}
All experiments are implemented in PyTorch on a RTX 4090 GPU. We use AdamW optimizer with initial learning rate of 1e-4 and maximum training epoch set at 100. We sampled SAR-RARP videos at 5Hz to get 40 videos with 46,586 normal frames and 18,540 error frames as the training set, and 8 videos with 7549 normal frames and 2096 error frames as the testing set. We use DINOv2 (giant version with registers) with fixed weights pretrained on ImageNet~\cite{dinov2} as the visual extractor to generate spatial embedding sequences $\mathcal{E}$ from the video with $L$ frames as the model input. The compression factor $G$ in the BMSS block is set as 64. The channel $E$ of dilated 1D convolutional layers in FCTF is fixed as $\frac{D}{8}$. Following ~\cite{mamba}, the state dimension $N$ of selective SSM is set to $16$. 

\subsection{Evaluation Metrics}
This work focuses on efficiently detecting errors in long surgical videos, formulating SED as a frame-level binary classification task. We therefore adopt the standard and widely used evaluation metrics--area under the curve (AUC) and average precision (AP)--at the frame level to assess performance. AUC and AP are both comprehensive metrics to measure the ability of a classifier to distinguish between positive (`\textit{error}') and negative (`\textit{normal}') classes. AUC refers to the area under the receiver operating characteristic (ROC) curve, which plots the true positive rate (TPR) against the false positive rate (FPR) across various threshold settings. Given a set of points on the ROC curve, where the points are ordered by increasing FPR, the AUC can be calculated using the trapezoidal rule:
\begin{equation}
\mathrm{AUC}=\sum_{i=1}^{n-1}\left(\frac{T P R_{i+1}+T P R_i}{2}\right) \cdot\left(F P R_{i+1}-F P R_i\right)
\end{equation}
where $T P R_i=\frac{T P_i}{T P_i+F N_i}$ and $F P R_i=\frac{F P_i}{F P_i+T N_i}$ are the TPR and FPR at point $i$, respectively. TP, TN, FP, and FN represent  the number of true positives, true negatives, false positives, and false negatives, respectively. $n$ is the number of thresholds. 

AP is calculated as the sum of the precision values weighted by the change in recall:
\begin{equation}
\mathrm{AP}=\sum_{i=1}^n\left(R_i-R_{i-1}\right) \cdot P_i
\end{equation}
where $R_i=\frac{TP_i}{TP_i+FN_i}$ and $P_i=\frac{TP_i}{TP_i+FP_i}$ are the recall and precision at the $i$-th threshold. Larger AUC and AP values indicate better performance. We report the mean and standard deviation from three runs with different random seeds.

\subsection{Comparison with the State-of-the-Art}
We compare the proposed SEDMamba with state-of-the-art SED method: LSTM~\cite{run_detection}; convolutional and Transformer-based video analysis methods: TeCNO~\cite{tecno}, MS-TCN~\cite{mstcn}, MS-TCN++~\cite{mstcn++}, ASFormer~\cite{asformer}; and selective SSM-based methods: Mamba~\cite{mamba} and Vim~\cite{zhu2024vision}. All methods are implemented based on their released code and original literature, and fine-tuned to fit our error detection task to provide a basis for a fair comparison. Note that we use an original Mamba/Vim block and a 1D convolutional layer with 1 kernel size and 1 channel to adapt Mamba~\cite{mamba}/Vim~\cite{zhu2024vision} to the SED task.

Table~\ref{tab_quan} shows quantitative results of all methods. For the overall performance (i.e. all test data regardless of error duration), the proposed SEDMamba consistently achieves superior performance with an increase of 1.82-13.06\% in AUC and 3.8-18.86\% in AP, compared to other methods. Notably, all selective SSM-based methods surpass the Transformer-based method ASFormer, demonstrating the superiority of selective SSM in the SED task. SEDMamba also outperforms other methods when grouping annotations together and detecting error instances instead of individual frames. Moreover, SEDMamba achieves gains of 2.19-18.68\% in AUC and 3.79-21.76\% in AP for short errors ($<$ 3 seconds), and 1.6-11.02\% in AUC and 1.61-13.78\% in AP for long errors ($\geq$ 3 seconds).

\begin{figure*}[!tbh]
    \centering
    \includegraphics[width=0.91\textwidth]{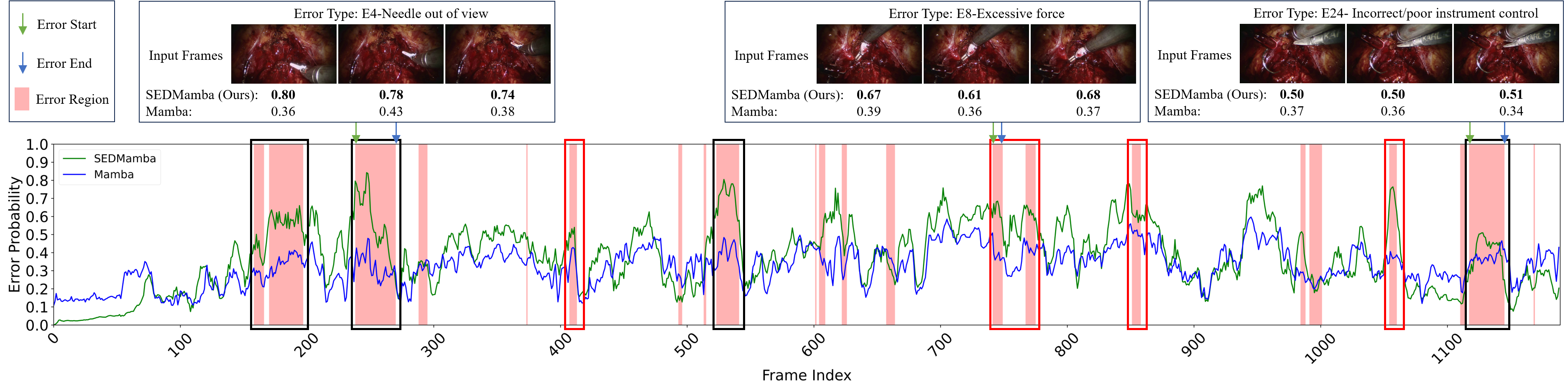}
    \includegraphics[width=0.91\textwidth]{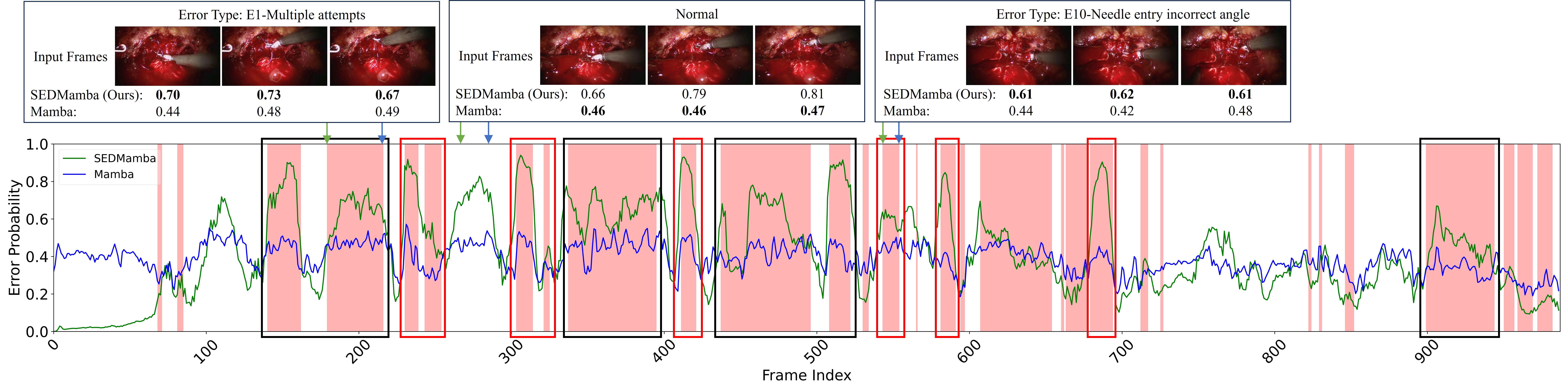}
    \includegraphics[width=0.91\textwidth]{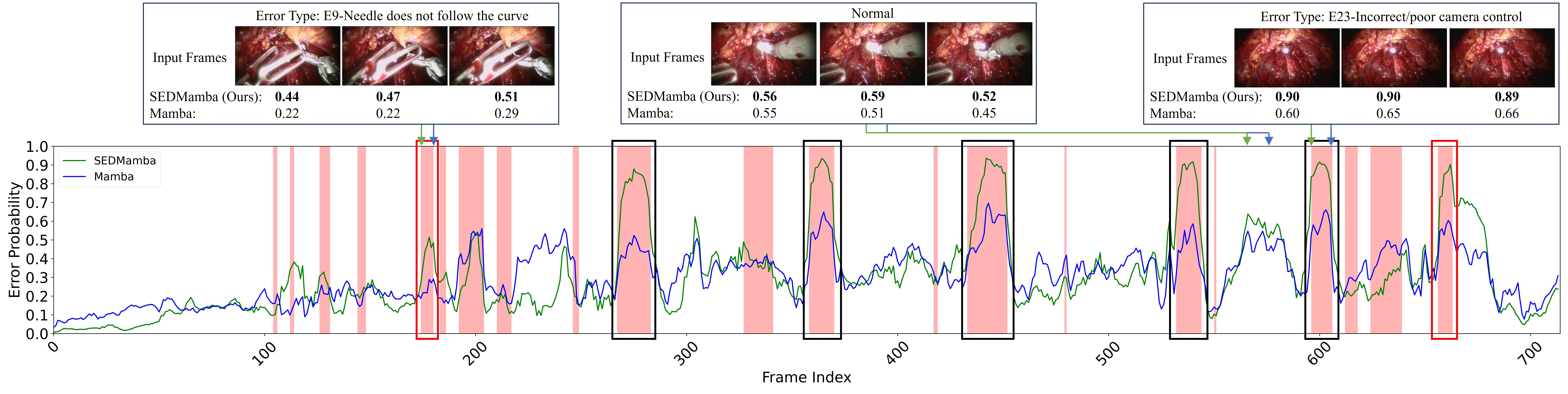}
    \includegraphics[width=0.91\textwidth]{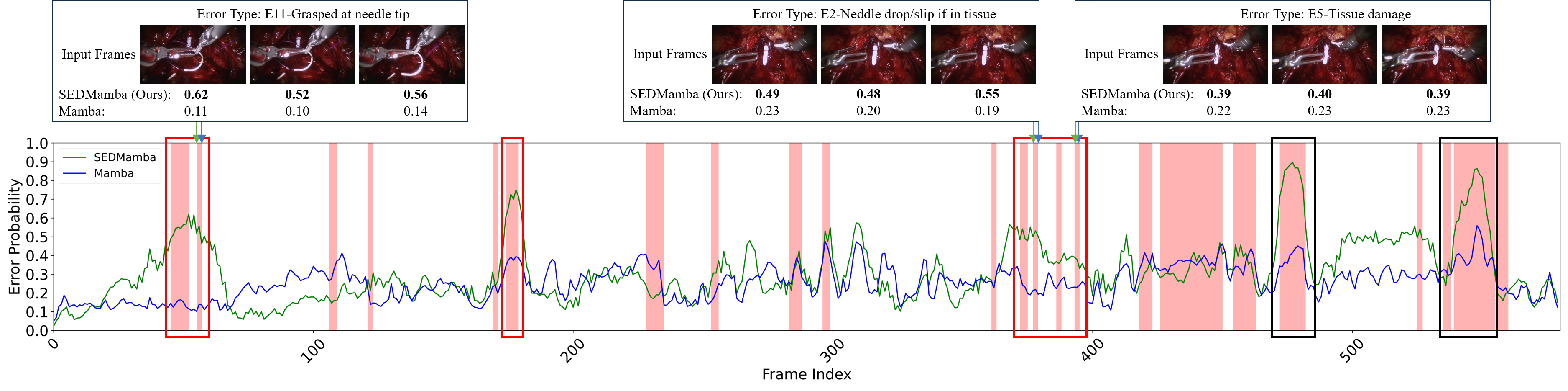}
    \caption{Visualization results of the proposed SEDMamba (green line) and the second-best method Mamba (blue line). Each row represents one complete test video. The X-axis is the frame index while the Y-axis is the error probability output by the models. The \colorbox{redtrans}{red translucent} background indicates error frames, while the white background indicates normal frames, i.e. ground truth of each frame. Different error types are considered, along with the corresponding prediction results (i.e., error probability). Red and black boxes are used to highlight significant results.}
    \label{fig_vis}
\end{figure*}

Fig.~\ref{fig_vis} illustrates typical error detection results on test videos.\footnote{Additional detection results via various error duration are included in the supplementary video.} The second-best method Mamba tends to output smoother and more ambiguous prediction curves, which impedes its sensitivity to errors. It also produces lower confidence outputs within error frames. Thanks to the fine-to-coarse temporal information provided by FCTF for selective SSM, SEDMamba is able to identify suddenly occurring errors with short duration (highlighted in red boxes) and consistently produces high confidence within error areas in most cases (highlighted in black boxes). In most normal frames, SEDMamba has comparable or lower error probabilities to Mamba, with only a small portion showing higher error probabilities. Specifically, when video quality deteriorates, such as instruments being covered with blood (second example in the 2$^{nd}$ row of Fig.~\ref{fig_vis}) or affected by glare (second example in the 3$^{rd}$ row of Fig.~\ref{fig_vis}), both SEDMamba and Mamba may produce false positives, with SEDMamba tending to output higher error probabilities. Note that both AUC and AP comprehensively consider all classes, i.e. `\textit{error}' and `\textit{normal}', indicating SEDMamba has superior performance in both classes.

\begin{table*}[!t]
\centering
\caption{Quantitative results of SEDMamba and other methods for all test data and different error duration ranges. We calculate AUC and AP at individual frame-level and per instance of error (AUC$_{ins}$ and AP$_{ins}$). Instance-level calculations are done by grouping consecutive frames with the same label as an error instance and then averaging the predicted error probabilities, across all frames within an instance, to obtain the overall error probability prediction per instance.}
\label{tab_quan}
\begin{tabular}{l|cccc|cc|cc}
\hline
\multirow{2}{*}{Methods}    & \multicolumn{4}{c|}{All Test Data} & \multicolumn{2}{c|}{Short Errors ($<$ 3 seconds)}                & \multicolumn{2}{c}{Long Errors ($\geq$ 3 seconds)}                 \\ \cline{2-9}             
& AUC (\%) $\uparrow$     & AP (\%) $\uparrow$  & AUC$_{ins}$ (\%) $\uparrow$     & AP$_{ins}$ (\%) $\uparrow$  & AUC (\%) $\uparrow$     & AP (\%) $\uparrow$  & AUC (\%) $\uparrow$     & AP (\%) $\uparrow$      \\ \hline
LSTM~\cite{run_detection}   &68.43$\pm$1.21    &37.26$\pm$2.65  &56.58$\pm$0.83  &60.35$\pm$0.93   &74.36$\pm$0.90   &33.06$\pm$1.27   &61.54$\pm$0.62   &21.34$\pm$1.56  \\
TeCNO~\cite{tecno}                 &58.14$\pm$0.33 &26.01$\pm$0.32  &51.87$\pm$0.56    &52.06$\pm$0.61  &59.24$\pm$0.36   &17.15$\pm$0.44   &57.65$\pm$0.85   &14.06$\pm$0.37 \\
MS-TCN~\cite{mstcn}                &58.82$\pm$0.16 &29.98$\pm$3.06  &51.17$\pm$0.86  &50.36$\pm$1.69    &62.50$\pm$1.23   &22.76$\pm$5.69    &54.70$\pm$1.68   &14.37$\pm$0.46 \\
MS-TCN++~\cite{mstcn++}              &58.57$\pm$0.43 &30.35$\pm$1.35    &51.69$\pm$0.12    &51.22$\pm$0.19  &63.57$\pm$0.87   &24.75$\pm$1.85   &52.99$\pm$0.06   &12.93$\pm$0.46 \\
ASFormer~\cite{asformer}              &68.76$\pm$0.93 &36.79$\pm$2.35   &54.92$\pm$0.87    &56.38$\pm$2.17  &74.77$\pm$2.02   &25.59$\pm$9.12   &62.66$\pm$2.74   &17.99$\pm$2.46 \\
Mamba~\cite{mamba}  &\underline{69.38}$\pm$0.26   &\underline{41.07}$\pm$0.95   &57.21$\pm$2.14    &\underline{62.24}$\pm$1.45  &\underline{75.73}$\pm$0.42   &\underline{35.12}$\pm$1.72   &\underline{62.41}$\pm$0.74   &\underline{25.10}$\pm$1.78   \\ 
Vim~\cite{zhu2024vision}   &69.06$\pm$1.46   &40.63$\pm$2.30    &\underline{57.76}$\pm$1.25    &62.02$\pm$2.53  &75.24$\pm$0.88   &34.64$\pm$1.54   &62.02$\pm$2.30   &24.67$\pm$1.08   \\ \hline
SEDMamba (Ours) &\textbf{71.20}$\pm$0.26 &\textbf{44.87}$\pm$1.52   &\textbf{59.28}$\pm$0.36    &\textbf{63.95}$\pm$0.57    &\textbf{77.92}$\pm$0.76   &\textbf{38.91}$\pm$0.57   &\textbf{64.01}$\pm$0.85   &\textbf{26.71}$\pm$0.50\\ \hline
\end{tabular}
\end{table*}

\subsection{Computational Efficiency} 
Given a video spatial embedding sequence $\mathcal{E} \in \mathcal{R}^{L\times D}$, the computational complexity of a single-head self-attention Transformer and a selective SSM in SEDMamba are $O(L^2D+LD^2)$ and $O(LDN)$, respectively. Computationally, self-attention in Transformer is quadratic to sequence length $L$, while selective SSM is linear to $L$. This computational efficiency makes SEDMamba scalable for gigapixel applications with large video lengths. Besides, as shown in Table~\ref{tab_quan} and Table~\ref{tab_cc}, SEDMamba achieves the highest performance yet maintains the lowest parameters and FLOPs. Specifically, SEDMamba achieves a 91.67\% reduction in parameters and a 92.32\% reduction in FLOPs compared to the cutting-edge SED method (LSTM), while delivering an AUC improvement of 2.77\% and an AP increase of 7.61\%. When compared to methods based on 1D and dilated convolution, i.e. TeCNO, MS-TCN and MS-TCN++, our method gets far more performance improvement with a smaller complexity. Furthermore, the parameters and FLOPs of our SEDMamba are only 27.38\% and 9.07\% of those of the Transformer-based method ASFormer, respectively, yet it achieves a performance gain of 2.44\% in AUC and 8.08\% in AP. Against other selective SSM-based methods, Mamba and Vim, our SEDMamba obtains 58.23\% and 75.79\% model parameter reductions, and 51.79\% and 70.90\% FLOPs reductions, respectively. This is attributed to the bottleneck construction of BMSS blocks, which compresses the spatial dimension at the input and restores it at the output for computational saving, as well as the FCTF therein BMSS, which provides fine-to-coarse temporal features for selective SSM to model long-term dependencies effectively.

\begin{table}[!t]
\centering
\caption{Computational complexity of SEDMamba and other methods in terms of model parameters (Params) and floating point operations (FLOPs). Lower Params and FLOPs mean lower model computational complexity. K = $\times 10^3$, M = $\times 10^6$.}
\label{tab_cc}
\begin{tabular}{l|cc}
\hline
Methods &Params (K) $\downarrow$ &FLOPs (M) $\downarrow$ \\ \hline
LSTM~\cite{run_detection}   &3481.79    &348.74 \\
TeCNO~\cite{tecno}                 &\underline{328.51} &\underline{45.17} \\
MS-TCN~\cite{mstcn}                &725.19 &71.98  \\
MS-TCN++~\cite{mstcn++}              &922.63  &91.64 \\
ASFormer~\cite{asformer}              &1059.27 &295.37 \\
Mamba~\cite{mamba}  &694.27  &55.55    \\ 
Vim~\cite{zhu2024vision}   &1197.83 &92.03   \\ \hline
SEDMamba (Ours) &\textbf{290.03} &\textbf{26.78} \\ \hline
\end{tabular}
\end{table}

\begin{table}[!t]
    \centering
    \caption{Ablation results for bottleneck mechanism and FCTF in BMSS blocks.}
    \label{tab_ab}
    \resizebox{\columnwidth}{!}{%
    \begin{tabular}{cc|cccc}
        \hline
        Bottleneck    &FCTF &Params (K) $\downarrow$   &FLOPs (M) $\downarrow$   &AUC  (\%) $\uparrow$   &AP (\%) $\uparrow$   \\ \hline
        &   &1569.79   &107.05   &67.98$\pm$1.68   &37.60$\pm$1.80   \\
        \checkmark   &    &267.11     &22.66   &\underline{69.62}$\pm$0.65    &\underline{41.29}$\pm$0.84   \\ 
        \checkmark &\checkmark   &290.03 &26.78    &\textbf{71.20}$\pm$0.26 &\textbf{44.87}$\pm$1.52  \\
        \hline
        \end{tabular}
        }
\end{table}

\begin{table}[!t]
\centering
\caption{Comparison of different numbers of BMSS blocks.}
\label{tab_ab_BMSS}
\begin{tabular}{c|cccc}
\hline
Blocks              & Params (K) $\downarrow$   &FLOPs (M) $\downarrow$   & AUC (\%) $\uparrow$     & AP (\%) $\uparrow$      \\ \hline
1        &137.35 &19.29         &69.96$\pm$1.29 &41.47$\pm$2.29 \\
2       &225.69 &30.09         &\underline{70.44}$\pm$0.49 &\underline{43.23}$\pm$0.78 \\
3   &290.03 &26.78           &\textbf{71.20}$\pm$0.26 &\textbf{44.87}$\pm$1.52 \\
4  &342.34 &41.16            &66.44$\pm$1.44 &38.01$\pm$0.80 \\
5  &388.64 &45.60   &66.21$\pm$2.76 &36.32$\pm$2.73 \\ \hline
\end{tabular}
\end{table}

\begin{table}[!t]
\centering
\caption{Comparison of different compression factor $G$ in BMSS blocks.}
\label{tab_cf}
\begin{tabular}{c|cccc}
\hline
G              & Params (K) $\downarrow$  &FLOPs (M) $\downarrow$   & AUC (\%) $\uparrow$     & AP (\%) $\uparrow$      \\ \hline
16        &54.61 &7.45         &67.60$\pm$1.74 &37.57$\pm$5.02 \\
32       &120.65 &12.99         &68.32$\pm$0.85 &39.80$\pm$2.46 \\
64   &290.03 &26.78           &\underline{71.20}$\pm$0.26 &\underline{44.87}$\pm$1.52 \\
128  &777.98 &65.12            &\textbf{71.84}$\pm$0.87 &\textbf{45.17}$\pm$1.47 \\ 
\hline
\end{tabular}
\end{table}

\begin{table}[!t]
\centering
\caption{Comparison of different numbers of dilated convolution layers in FCTF.}
\label{tab_ab_FCTF}
\resizebox{\columnwidth}{!}{%
\begin{tabular}{l|cccc}
\hline
Numbers [Dilation Rates]              & Params (K) $\downarrow$   &FLOPs (M) $\downarrow$   & AUC (\%) $\uparrow$     & AP (\%) $\uparrow$      \\ \hline
0    &267.11     &22.66   &69.62$\pm$0.65    &41.29$\pm$0.84 \\
1 [2]        &285.76 &26.35         &69.85$\pm$0.74 &41.37$\pm$0.44 \\
2 [2, 4]      &287.90 &26.57          &69.89$\pm$0.56 &41.77$\pm$0.80\\
3 [2, 4, 8]  &290.03 &26.78           &\textbf{71.20}$\pm$0.26 &\textbf{44.87}$\pm$1.52 \\
4 [2, 4, 8, 16] &292.17 &26.99            &\underline{70.45}$\pm$0.90 &\underline{43.36}$\pm$2.50 \\
5 [2, 4, 8, 16, 32] &294.30 &27.20   &70.30$\pm$1.57 &42.33$\pm$1.98 \\ \hline
\end{tabular}
}
\end{table}

\begin{table}[!t]
\centering
\caption{Comparison of Mamba alternatives in BMSS blocks.}
\label{tab:alt}
\resizebox{\columnwidth}{!}{
\begin{tabular}{l|cccc}
\hline
Methods              & Params (K) $\downarrow$   &FLOPs (M) $\downarrow$   & AUC (\%) $\uparrow$     & AP (\%) $\uparrow$      \\ \hline
BMSS$_{LSTM}$    &291.95     &29.22   &68.78$\pm$0.06    &41.58$\pm$1.65 \\
BMSS$_{TeCNO}$        &600.88 &19.08         &67.19$\pm$0.26 &39.84$\pm$0.23 \\
BMSS$_{ASFormer}$      &354.55 &35.28          &66.83$\pm$1.03 &34.70$\pm$1.88  \\ 
BMSS (our SEDMamba)         &\textbf{290.03} &\textbf{26.78}    &\textbf{71.20}$\pm$0.26 &\textbf{44.87}$\pm$1.52 \\ \hline
\end{tabular}
}
\end{table}

\subsection{Ablation Study}
\subsubsection{Effectiveness of Key Components in BMSS Block}
We study the effect of bottleneck mechanism and FCTF in BMSS blocks. As shown in Table~\ref{tab_ab}, whereby removing each component leads to performance degradation. The bottleneck mechanism yields performance gains of +1.64\% in AUC and +3.69\% in AP, while model parameters and FLOPs are significantly decreased by 82.98\% and 78.83\%, which guards against overfitting and enhances model efficiency. Besides, such a bottleneck mechanism compresses features on the spatial dimension, which can mitigate the impact of redundant spatial noise on long-term dependency modelling. Introducing FCTF to bottleneck mechanism leads to a slight increase in computational complexity, which is acceptable given the substantial performance gains of +1.58\% in AUC and +3.58\% in AP. This underscores the importance of providing multi-scale temporal information for selective SSM in the SED task.

\subsubsection{Number of BMSS Blocks}
As shown in Table~\ref{tab_ab_BMSS}, the depth of our SEDMamba can be tailored by hierarchically stacking BMSS blocks. Remarkably, even with a single BMSS block, our method achieves an AUC of 69.96\% and an AP of 41.47\%, surpassing the second-best method Mamba, while our method's parameters and FLOPs are only 19.78\% and 34.73\% of Mamba's. This further verifies the effectiveness of the proposed BMSS block. Performance continues to improve with more BMSS blocks but declines when the model becomes too deep, such as with five BMSS blocks. The potential reason behind this decrease could be the gradient vanishing/exploding issue commonly found in deep models. Introducing residual connections between BMSS blocks might be helpful.

\subsubsection{Compression Factor in Bottleneck Mechanism}
The compression factor $G$ of the bottleneck mechanism in the BMSS block determines the spatial dimension of the input for the FCTF and selective SSM. As shown in Table~\ref{tab_cf}, it is evident that altering $G$ significantly affects the computational complexity of the model. Considering the trade-off between performance and computational complexity, we set $G$ to 64.

\subsubsection{Number of dilated convolutional layers in FCTF}
As shown in Table~\ref{tab_ab_FCTF}, incorporating progressively coarser temporal information via dilated convolutions generally improves performance, peaking at three layers ([2, 4, 8] dilation rates). Beyond this, performance gains begin to decline, likely due to the introduction of redundant coarse-grained features that complicate feature fusion and reduce the model's ability to focus on salient temporal information. These findings suggest that a balanced inclusion of dilated convolutions should be considered for effectively capturing coarse temporal information.

\subsubsection{Alternatives to the Mamba part in BMSS}
To explore the impact of the Mamba part (i.e., the part between FCTF and the final 1D convolution layer in BMSS), we substitute it with other key temporal components. Specifically, we replace Mamba in BMSS with the LSTM layer from \cite{run_detection}, the TCN block from \cite{tecno}, and encoder-decoder blocks from \cite{asformer}, resulting in three BMSS variants named BMSS$_{LSTM}$, BMSS$_{TeCNO}$, and BMSS$_{ASFormer}$. As shown in Table~\ref{tab:alt}, SEDMamba achieves the best performance with lowest computational complexity, indicating the effectiveness of Mamba. We also observed that the bottleneck and FCTF further unleash the potential of the TCN block and LSTM layer, improving both computational efficiency and performance.

\subsubsection{Experiments with the JIGSAWS dataset}
To explore the generalizability of our proposed SEDMamba method to other surgical scenes, we extend it to the JIGSAWS dataset with error annotations from \cite{jigsaws_errors}. Following prior work~\cite{10750058}, which utilized video data from the JIGSAWS dataset for surgical error detection, our method obtains 75.41$\pm$2.27\% AUC and 77.76$\pm$5.04\% AP under the Leave-One-Supertrial-Out (LOSO) setting. Compared to Mamba and Vim, SEDMamba gains 0.91-1.34\% in AUC and 0.65-1.23\% in AP, demonstrating its feasibility in dry-lab scenes.

\section{Conclusion}
In this paper, we build a novel SED dataset by providing the first-of-its-kind, frame-level error annotations of in-vivo RAS cases. More than 970000 frames from 48 videos in SAR-RARP50 are annotated to provide temporal boundaries of surgical errors. We then propose SEDMamba, a hierarchical SED model based on selective SSM with bottleneck and fine-to-coarse temporal fusion, which yields significant improvements in error detection performance with high computational efficiency. Experimental results show that the proposed method outperforms seven state-of-the-art competitions with 1.82-13.06\% AUC and 3.8-18.86\% AP performance improvements and better qualitative results. Exploiting a very efficient design SEDMamba requires between 11.71-91.67\% less model parameters leading to a 40.71-92.32\% FLOPs reduction. Future work will focus on detecting error types semantically with multi-label algorithms.

\bibliographystyle{IEEEtran}
\bibliography{refs}

\begin{thebibliography}{10}
\providecommand{\url}[1]{#1}
\csname url@rmstyle\endcsname
\providecommand{\newblock}{\relax}
\providecommand{\bibinfo}[2]{#2}
\providecommand\BIBentrySTDinterwordspacing{\spaceskip=0pt\relax}
\providecommand\BIBentryALTinterwordstretchfactor{4}
\providecommand\BIBentryALTinterwordspacing{\spaceskip=\fontdimen2\font plus
\BIBentryALTinterwordstretchfactor\fontdimen3\font minus \fontdimen4\font\relax}
\providecommand\BIBforeignlanguage[2]{{%
\expandafter\ifx\csname l@#1\endcsname\relax
\typeout{** WARNING: IEEEtran.bst: No hyphenation pattern has been}%
\typeout{** loaded for the language `#1'. Using the pattern for}%
\typeout{** the default language instead.}%
\else
\language=\csname l@#1\endcsname
\fi
#2}}

\bibitem{robot_rarp}
Y.~Du, \emph{et~al.}, ``Robot-assisted radical prostatectomy is more beneficial for prostate cancer patients: A system review and meta-analysis,'' \emph{Med. Sci. Monit.}, vol.~14, no.~24, pp. 272--287, 2018.

\bibitem{sarker_errors}
S.~K. Sarker and C.~Vincent, ``Errors in surgery,'' \emph{Int. Surg. J.}, vol.~3, no.~1, pp. 75--81, 2005.

\bibitem{alemzadeh_adverse}
H.~Alemzadeh, J.~Raman, N.~Leveson, Z.~Kalbarczyk, and R.~K. Iyer, ``Adverse events in robotic surgery: A retrospective study of 14 years of fda data,'' \emph{PLoS One}, vol.~11, no.~4, p. e0151470, 2016.

\bibitem{tang_ochra}
B.~Tang and A.~Cuschieri, ``Objective assessment of surgical operative performance by observational clinical human reliability analysis (ochra): a systematic review,'' \emph{Surg. Endosc.}, vol.~34, no.~4, pp. 1492--1508, 2020.

\bibitem{gorard_ochra}
J.~Gorard, M.~Boal, V.~Swamynathan, W.~Ghamrawi, and N.~Francis, ``The application of objective clinical human reliability analysis (ochra) in the assessment of basic robotic surgical skills,'' \emph{Sur. Endosc.}, vol.~38, no.~4, pp. 116--128, 2024.

\bibitem{eubanks_score}
T.~R. Eubanks, \emph{et~al.}, ``An objective scoring system for laparoscopic cholecystectomy,'' \emph{J. Am. Coll. Surg.}, vol. 189, no.~6, pp. 566--574, 1999.

\bibitem{curtis_eval}
N.~J. Curtis, \emph{et~al.}, ``Clinical evaluation of intraoperative near misses in laparoscopic rectal cancer surgery,'' \emph{Ann. Surg.}, vol. 273, no.~4, pp. 778--784, 2021.

\bibitem{curtis_assoc}
N.~J. Curtis, \emph{et~al.}, ``Association of surgical skill assessment with clinical outcomes in cancer surgery,'' \emph{JAMA Surg.}, vol. 155, no.~7, pp. 590--598, 2020.

\bibitem{boal_review}
M.~W. Boal, \emph{et~al.}, ``Evaluation of objective tools and artificial intelligence in robotic surgery technical skills assessment: a systematic review,'' \emph{Br. J. Surg.}, vol. 111, no.~1, p. znad331, 2024.

\bibitem{sds}
L.~Maier-Hein, \emph{et~al.}, ``Surgical data science for next-generation interventions,'' \emph{Nat. Biomed. Eng.}, vol.~1, no.~9, pp. 691--696, 2017.

\bibitem{jigsaws}
Y.~Gao, \emph{et~al.}, ``The jhu-isi gesture and skill assessment working set (jigsaws): A surgical activity dataset for human motion modeling,'' in \emph{Proc. Med. Image. Comput. Comput. Assist. Interv. Workshop}, 2014.

\bibitem{funke_3d}
I.~Funke, S.~Bodenstedt, F.~Oehme, F.~von Bechtolsheim, J.~Weitz, and S.~Speidel, ``Using 3d convolutional neural networks to learn spatiotemporal features for automatic surgical gesture recognition in video,'' in \emph{Proc. Med. Image. Comput. Comput. Assist. Interv.}, 2019.

\bibitem{zhang_symm}
J.~Zhang, \emph{et~al.}, ``Symmetric dilated convolution for surgical gesture recognition,'' in \emph{Proc. Med. Image. Comput. Comput. Assist. Interv.}, 2020.

\bibitem{liu_deep}
D.~Liu and T.~Jiang, ``Deep reinforcement learning for surgical gesture segmentation and classification,'' in \emph{Proc. Med. Image. Comput. Comput. Assist. Interv.}, 2018.

\bibitem{dipietro_reco}
R.~DiPietro, \emph{et~al.}, ``Recognizing surgical activities with recurrent neural networks,'' in \emph{Proc. Med. Image. Comput. Comput. Assist. Interv.}, 2016.

\bibitem{zhang_sdnet}
J.~Zhang, Y.~Nie, Y.~Lyu, X.~Yang, J.~Chang, and J.~J. Zhang, ``Sd-net: joint surgical gesture recognition and skill assessment,'' \emph{Int. J. CARS}, vol.~16, pp. 1675--1682, 2021.

\bibitem{act_review}
B.~van Amsterdam, M.~J. Clarkson, and D.~Stoyanov, ``Gesture recognition in robotic surgery: A review,'' \emph{IEEE Trans. Biomed. Eng.}, vol.~68, no.~6, pp. 2021--2035, 2021.

\bibitem{rarp}
B.~Van~Amsterdam, \emph{et~al.}, ``Gesture recognition in robotic surgery with multimodal attention,'' \emph{IEEE Trans. Med. Imaging}, vol.~41, no.~7, pp. 1677--1687, 2022.

\bibitem{kiyasseh_vis}
D.~Kiyasseh, \emph{et~al.}, ``A vision transformer for decoding surgeon activity from surgical videos,'' \emph{Nat. Biomed. Eng}, vol.~7, pp. 780--796, 2023.

\bibitem{ma_gest}
R.~Ma, \emph{et~al.}, ``Surgical gestures as a method to quantify surgical performance and predict patient outcomes,'' \emph{npj Digit. Med.}, vol.~5, no. 187, 2022.

\bibitem{jigsaws_errors}
K.~Hutchinson, Z.~Li, L.~A. Cantrell, N.~S. Schenkman, and H.~Alemzadeh, ``Analysis of executional and procedural errors in dry-lab robotic surgery experiments,'' \emph{Int. J. Med. Robot.}, vol.~18, no.~3, p. e2375, 2022.

\bibitem{run_detection}
Z.~Li, K.~Hutchinson, and H.~Alemzadeh, ``Runtime detection of executional errors in robot-assisted surgery,'' in \emph{Proc. IEEE Int. Conf. Robotics Autom.}, 2022.

\bibitem{drautoencoder}
D.~J. Samuel and F.~Cuzzolin, ``Unsupervised anomaly detection for a smart autonomous robotic assistant surgeon (saras) using a deep residual autoencoder,'' \emph{IEEE Robot. Autom. Lett.}, vol.~6, no.~4, pp. 7256--7261, 2021.

\bibitem{asformer}
F.~Yi, H.~Wen, and T.~Jiang, ``Asformer: Transformer for action segmentation,'' in \emph{Proc. Br. Mach. Vis. Conf.}, 2021.

\bibitem{transsv}
X.~Gao, Y.~Jin, Y.~Long, Q.~Dou, and P.-A. Heng, ``Trans-svnet: Accurate phase recognition from surgical videos via hybrid embedding aggregation transformer,'' in \emph{Proc. Med. Image. Comput. Comput. Assist. Interv.}, 2021, pp. 593--603.

\bibitem{s4}
A.~Gu, K.~Goel, and C.~R{\'e}, ``Efficiently modeling long sequences with structured state spaces,'' in \emph{Proc. Int. Conf. Learn. Represent.}, 2021.

\bibitem{mamba}
A.~Gu and T.~Dao, ``Mamba: Linear-time sequence modeling with selective state spaces,'' \emph{arXiv preprint arXiv:2312.00752}, 2023.

\bibitem{zhu2024vision}
L.~Zhu, B.~Liao, Q.~Zhang, X.~Wang, W.~Liu, and X.~Wang, ``Vision mamba: Efficient visual representation learning with bidirectional state space model,'' in \emph{Proc. Int. Conf. Mach. Learn.}, 2024.

\bibitem{xing2024segmamba}
Z.~Xing, T.~Ye, Y.~Yang, G.~Liu, and L.~Zhu, ``Segmamba: Long-range sequential modeling mamba for 3d medical image segmentation,'' \emph{arXiv preprint arXiv:2401.13560}, 2024.

\bibitem{sarrarp50}
D.~Psychogyios, \emph{et~al.}, ``Sar-rarp50: Segmentation of surgical instrumentation and action recognition on robot-assisted radical prostatectomy challenge,'' \emph{arXiv preprint arXiv:2401.00496}, 2024.

\bibitem{rarp45}
B.~van Amsterdam, \emph{et~al.}, ``Gesture recognition in robotic surgery with multimodal attention,'' \emph{IEEE Trans. Med. Imaging}, vol.~41, no.~7, pp. 1677--1687, 2022.

\bibitem{guni_check}
A.~Guni, N.~Raison, B.~Challacombe, S.~Khan, P.~Dasgupta, and K.~Ahmed, ``Development of a technical checklist for the assessment of suturing in robotic surgery,'' \emph{Surg. Endosc.}, vol.~32, pp. 4402---4407, 2018.

\bibitem{dinov2}
M.~Oquab, \emph{et~al.}, ``Dinov2: Learning robust visual features without supervision,'' \emph{arXiv preprint arXiv:2304.07193}, 2023.

\bibitem{elfwing2018sigmoid}
S.~Elfwing, E.~Uchibe, and K.~Doya, ``Sigmoid-weighted linear units for neural network function approximation in reinforcement learning,'' \emph{Neural Netw.}, vol. 107, pp. 3--11, 2018.

\bibitem{tecno}
T.~Czempiel, \emph{et~al.}, ``Tecno: Surgical phase recognition with multi-stage temporal convolutional networks,'' in \emph{Proc. Med. Image. Comput. Comput. Assist. Interv.}, 2020, pp. 343--352.

\bibitem{mstcn}
Y.~A. Farha and J.~Gall, ``Ms-tcn: Multi-stage temporal convolutional network for action segmentation,'' in \emph{Proc. IEEE/CVF Conf. Comput. Vis. Pattern Recognit.}, 2019, pp. 3575--3584.

\bibitem{mstcn++}
S.~Li, Y.~A. Farha, Y.~Liu, M.-M. Cheng, and J.~Gall, ``Ms-tcn++: Multi-stage temporal convolutional network for action segmentation,'' \emph{IEEE Trans. Pattern Anal. Mach. Intell.}, vol.~45, no.~6, pp. 6647--6658, 2020.

\bibitem{10750058}
Z.~Shao, J.~Xu, D.~Stoyanov, E.~B. Mazomenos, and Y.~Jin, ``Think step by step: Chain-of-gesture prompting for error detection in robotic surgical videos,'' \emph{IEEE Robot. Autom. Lett.}, vol.~9, no.~12, pp. 11\,513--11\,520, 2024.

\end{thebibliography}

\newpage

\vfill

\end{document}